\documentclass{llncs}
\usepackage{graphicx}
\usepackage{amsmath,amssymb} 
\usepackage{color}
\usepackage{subfig}
\usepackage{adjustbox}
\usepackage{multirow}
\usepackage[width=122mm,left=12mm,paperwidth=146mm,height=193mm,top=12mm,paperheight=217mm]{geometry}
\usepackage{ctable}
\newcommand{\etal}{\textit{et al}. }
\newcommand{\ie}{\textit{i}.\textit{e}., }

\DeclareMathOperator*{\argfind}{arg\,find}
\newcommand*{\affaddr}[1]{#1} 
\newcommand*{\affmark}[1][*]{\textsuperscript{#1}}
\begin{document}
\pagestyle{headings}
\mainmatter
\def\ECCV16SubNumber{*}  

\title{Progressive Neural Networks for Image Classification} 

\author{%
	Zhi~Zhang\affmark[1], Guanghan~Ning\affmark[1], Yigang~Cen\affmark[2], Yang~Li\affmark[1], Zhiqun~Zhao\affmark[1], Hao~Sun\affmark[1] and Zhihai~He\affmark[1]\\
	\affaddr{\affmark[1]Department of EECS, University of Missouri-Columbia}\\
	\affaddr{\affmark[2]Department of Computer Science, Beijing Jiaotong University}\\%
}
\institute{}

\maketitle

\begin{abstract}

The inference structures and computational complexity of existing deep neural networks, once trained, are fixed and remain the same for all test images. However, in practice, it is highly desirable to establish a progressive structure for deep neural networks which is able 
to adapt its inference process and complexity for images with different visual recognition complexity.  In this work, we develop a multi-stage progressive structure with integrated confidence analysis and decision policy learning for deep neural networks. This new framework consists of a set of network units to be  activated in a sequential manner with progressively increased complexity and visual recognition power. Our extensive experimental results on the CIFAR-10  and ImageNet datasets demonstrate that the proposed progressive deep neural network  is able to obtain more than 10 fold complexity scalability while achieving the state-of-the-art performance using a single network model satisfying different complexity-accuracy requirements.

\end{abstract}

\section{Introduction}

Recently, large and deep neural networks have demonstrated extraordinary performance on various computer vision and machine learning tasks \cite{ILSVRC15,huang2017densely}. 
We notice that, the inference structures, execution procedures, and computational complexity of existing deep neural networks,  once trained, are fixed and remain the same for all test images, no matter how much they have been optimized speed-wise. 
In this work, our goal is to develop a new progressive framework for deep neural network such that a single model can be evaluated at different performance levels with different computational complexity requirements. This single-model-variable-complexity property is very important in practice.  We recognize that different images have different complexity levels of visual representation and different difficulty levels for visual recognition. For simple images with low visual recognition complexity, we can easily classify the image or recognize the object with simple networks at very low complexity. 
For example, it will be very easy to detect a person standing in front of a clean background or classify if it is male or female. For harder images, we will need to extract sophisticated visual features using more layers of network representation  to gain sufficient visual discriminative power so that the object can be successfully distinguished from those in other classes.

\begin{figure}
\centering
\hspace{-0.5cm}\parbox{4.5cm}{
\includegraphics[width=1.1\linewidth]{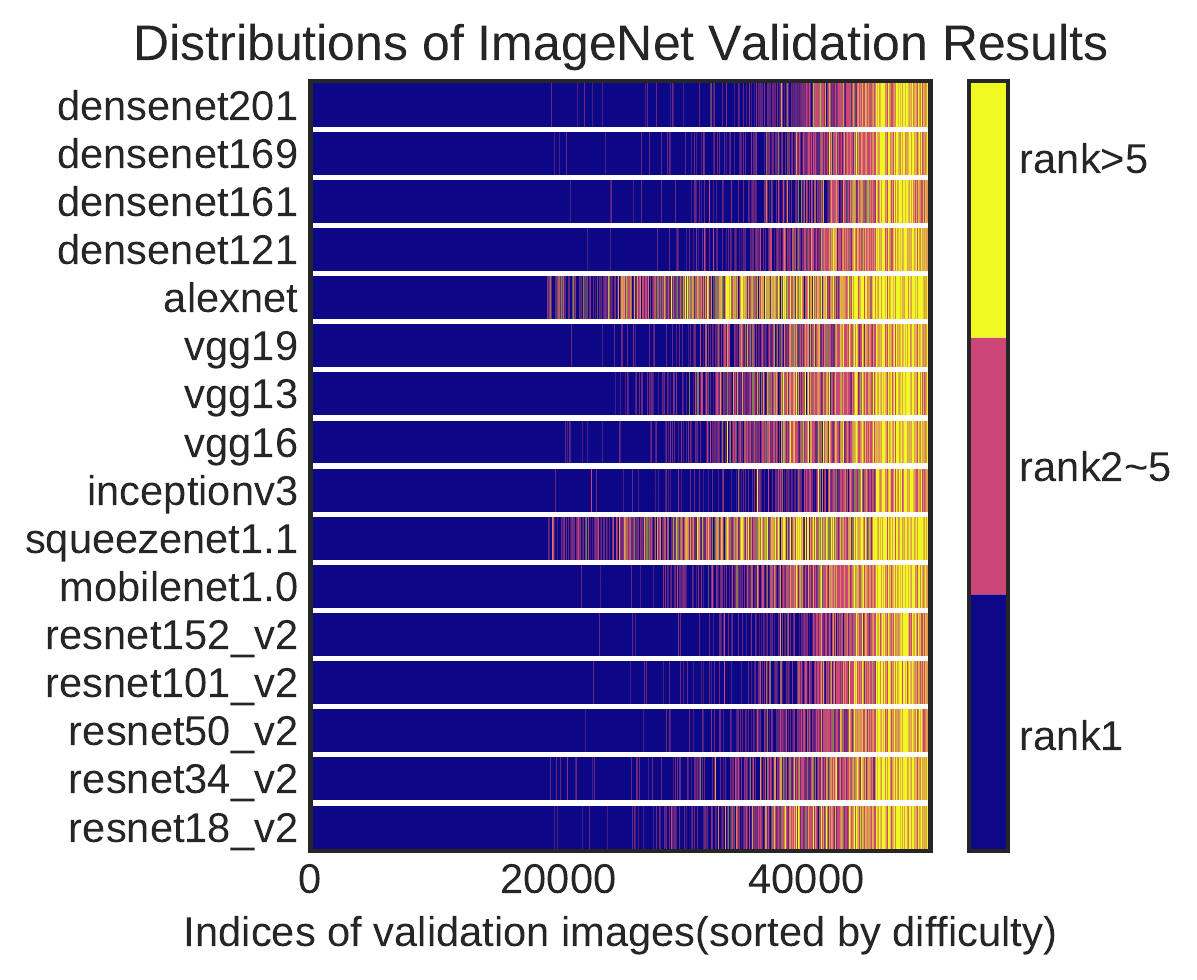}
\label{fig:valid_distribution}}
\qquad
\begin{minipage}{4.5cm}
\includegraphics[width=1 \linewidth]{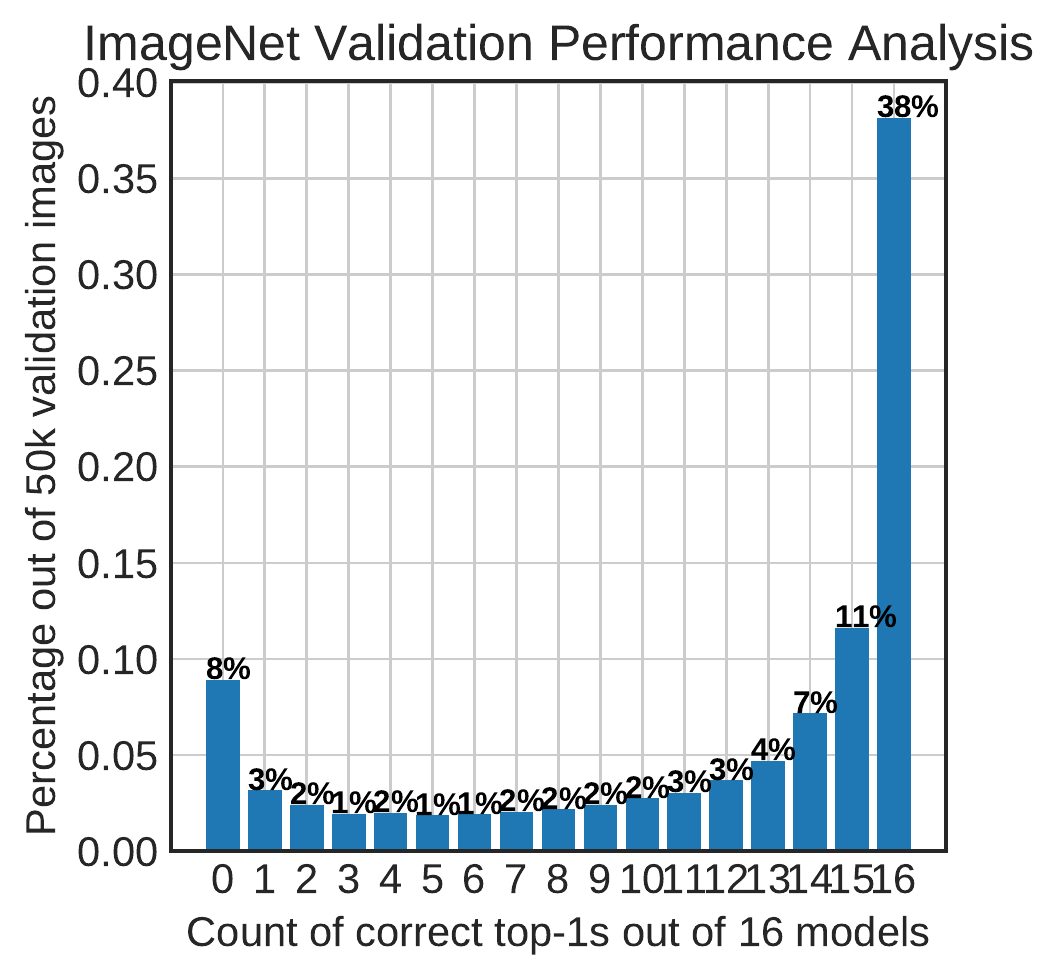}
\label{fig:valid_counts}
\end{minipage}
\vspace{-0.5cm}
\caption{\small \textbf{Left}: Validation results of different models, images are sorted by average top-1 accuracies across models. \textbf{Right}: Individual image difficulties across different models. }
\label{fig:distribution}
\end{figure}

To validate this observation, we conducted the following interesting experiment. We collected 16 different pre-trained deep neural networks, ranging from very low-complexity MobileNet \cite{howard2017mobilenets}, to very high-complexity DenseNet201 \cite{huang2017densely}.  We use these networks to classify the images in the ILSVRC2012 \cite{ILSVRC15} 50k validation set.
Fig. \ref{fig:distribution} (left) shows the classification results. Each row corresponds to a specific network. The horizontal axis represents the index of the test image. A blue line indicates that the image is successfully classified by the network or ranked first in all images. A magenta line indicates that the correct result is within the top 2 to 5 categories. A yellow line indicates that the correct result is beyond the top 5 results. As summarized in Fig. \ref{fig:distribution}, we can see that 38\% of images are always correctly classified by all networks, no matter how simple the network is. These are the so-called simple images with low visual recognition complexity. 
This suggests that, if we can successfully identify those set of simple images, we can save a lot of computational resources by choosing simple networks to analyze them. More excitingly, if we are able to model or predict the visual recognition complexity of the input image and if we are able to establish a progressive network, we can then adapt the network complexity during run-time according to the visual recognition complexity of the input image. This will allow us to save the computational resources significantly.

\begin{figure}
\centering
\includegraphics[width=0.99\linewidth]{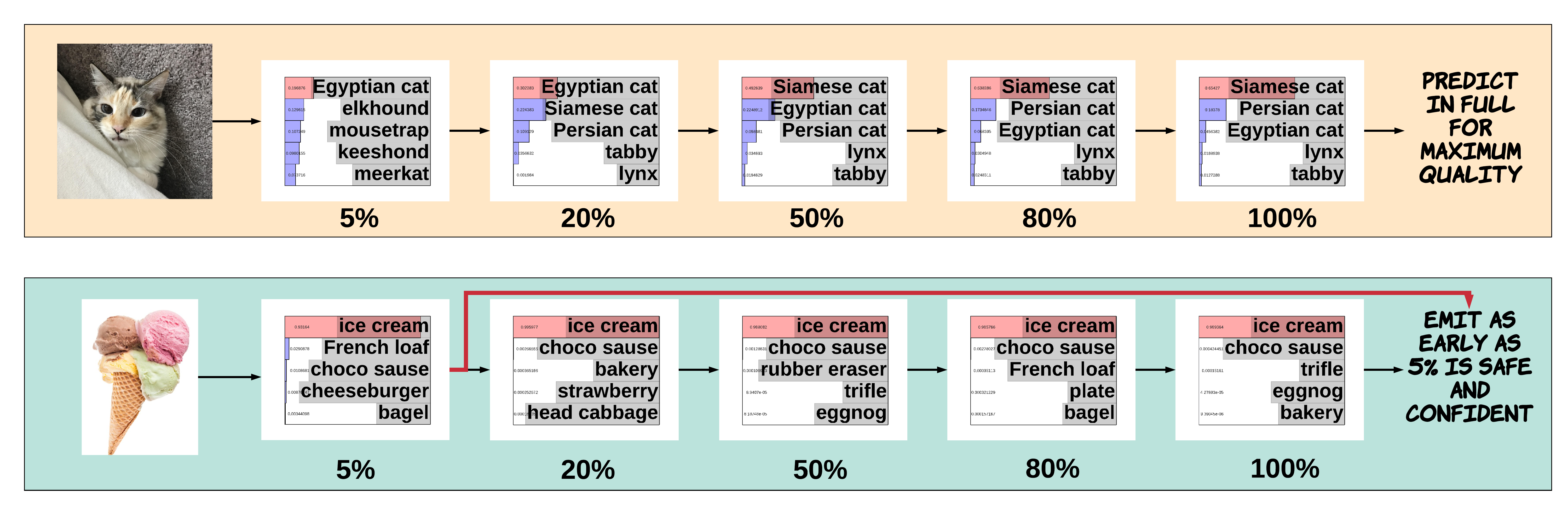}
\caption{\small  Each prediction result is produced by pre-trained model of different size.  Top row: hard example which requires full network inference. Bottom row: easy example that is predicted confidently by even tiny models. \textbf{Best viewed in color.}}
\label{fig:intuition}
\end{figure}

Let us look at one more example. Fig. \ref{fig:intuition} shows two images, a simple Ice Cream image and a hard Siamese Cat image with occlusion. We choose 5 networks with different computational complexity. The most complex network is the DenseNet201 \cite{huang2017densely} labeled with 100\%. The complexity levels of the other four networks are shown in approximate relative percentage in Fig. \ref{fig:intuition}. For example, the inference cost of the first network is about 5\% of DenseNet201.
From this experiment, we can see that, for the simple Ice Cream image, the confidence score for Ice Cream is much higher than other object categories, even with very simple networks. However, for the hard Siamese Cat image, the score distribution is more uniform for low-complexity networks. As the network becomes more complex and more powerful, the classifier is more and more confident about the result since the score for the correct label is getting much higher than other categories. 

These two experiments strongly suggest that  it is highly desirable to establish a progressive deep neural network structure 
which is able to adapt its network inference structure and computational complexity to images with different visual recognition complexity. This progressive structure should be able to scale up its discriminative power and achieve higher recognition capability by activating and executing more analysis units of the deep neural networks and accumulating more visual evidences for more challenging vision analysis tasks or image sets.  

The major contributions of  this work can be summarized as follows. (1) We have successfully developed a multi-stage progressive structure, called ProgNet,  for deep neural networks, with each stage being separately evaluated for output decision and all stages being  activated in a sequential manner with progressively increased complexity and visual recognition power.
(2) We present different structures for progress network design and develop a Confidence Analysis and Decision Policy network to learn the classification confidence function for the progressive network and make run-time complexity-accuracy decision for each input image.  
(3) Our extensive experimental results demonstrate that the proposed progressive framework for deep neural networks is able to outperform existing state-of-the-art networks or models, from MobileNet to DenseNet, using one single model and its complexity can be adaptively controlled. This progressive framework  will provide an important and useful tool for practical  deep neural network design and resource allocation in real-time applications. 

The rest of the paper is organized as follows. Section \ref{sec-related} reviews related work. Section \ref{sec-overview} provides a conceptual overview of the proposed progressive framework. Detailed design and methods for progressive networks are presented in Section \ref{sec:progressive}. Section \ref{sec-cadp} presents our confidence analysis and decision policy network. Experimental results are presented in Section \ref{sec:results}. 
Section \ref{sec:conclusion} concludes the paper.

\section{Related Work}
\label{sec-related}
This is work is closely related to complexity control / optimization 
and confidence analysis of deep neural networks.

\textbf{(A) Complexity Optimization of Deep Neural Networks.} Deep neural networks often involves high computational complexity. 
A number of methods have been developed to accelerate inference speed of deep neural networks, or reduce its computational resource requirement so that they can operate on 
lower-end devices, such as CPUs, embedded processors and mobile devices. Network parameter pruning and quantization are two effective approaches.  
Gong \etal \cite{gong2014compressing} and Wu \cite{wu2016quantized} applied k-means scalar quantization to pre-trained parameters. Significant speed up can be achieved by 8-bit integer and 16-bit floating point quantization as shown in \cite{vanhoucke2011improving} and \cite{gupta2015deep}. 
Parameter pruning approaches \cite{hanson1989comparing,lecun1990optimal,hassibi1993second} can be used to reduce network complexity. Low rank factorization and decomposition \cite{tai2015convolutional}, transferred learning and compact convolutional filter learning \cite{cohen2016group,shang2016understanding}, and knowledge distillation \cite{buciluǎ2006model,ba2014deep,hinton2015distilling} methods have been developed to reduce the network complexity. 
NoScope \cite{kang2017noscope} tackled the problem of very expensive surveillance video object detection by using a shallow and fast CNN as a early estimator and dispatcher. Only complex scenes with significant inter-frame changes will require a full run of deep object detection network, therefore higher analysis speed can be achieved.  
All of these networks aim to optimize the computational complexity and inference speed of deep neural networks, often at the cost of degraded visual recognition performance. Their inference structures, execution procedures, and computational complexity, once trained, are fixed and remain the same for all test images. They are not able to adapt their network inference structure and complexity to different resource supplies and input images.

\textbf{(B) Confidence Analysis for Deep Neural Networks.}
Researchers have recognized that the decision scores of existing deep neural networks are poorly calibrated \cite{guo2017calibration}. For example, higher scores often do not correspond to better or closer samples.   \cite{subramanya2017confidence} argues that probability scores generated by \textit{softmax} should not be considered as confidence score or distance measure directly since they are based on the $l_2$ norm of \textit{pre-softmax} inputs. To address this issue, various methods using scalar, vector, matrix and binning methods to calibrate the confidence scores produced by the \textit{softmax} function.  
Gaussian density modeling is proposed in \cite{subramanya2017confidence} as a post prediction calibration  using prior information of the training data. 
Gal \etal \cite{gal2016dropout} has implemented a randomized dropout network \cite{srivastava2014dropout}  to estimate the uncertainty level of network prediction. 
The open set deep network approach in \cite{bendale2016towards} attempts to measure uncertainty contributed from unknown categories. 
It should be noted that these methods are based on statistical modeling, being optimized on the entire validation set, and therefore not suitable for confidence analysis for each individual input image.

\section{Method Overview}
\label{sec-overview}
\begin{figure}
\centering
\includegraphics[width=0.8\linewidth]{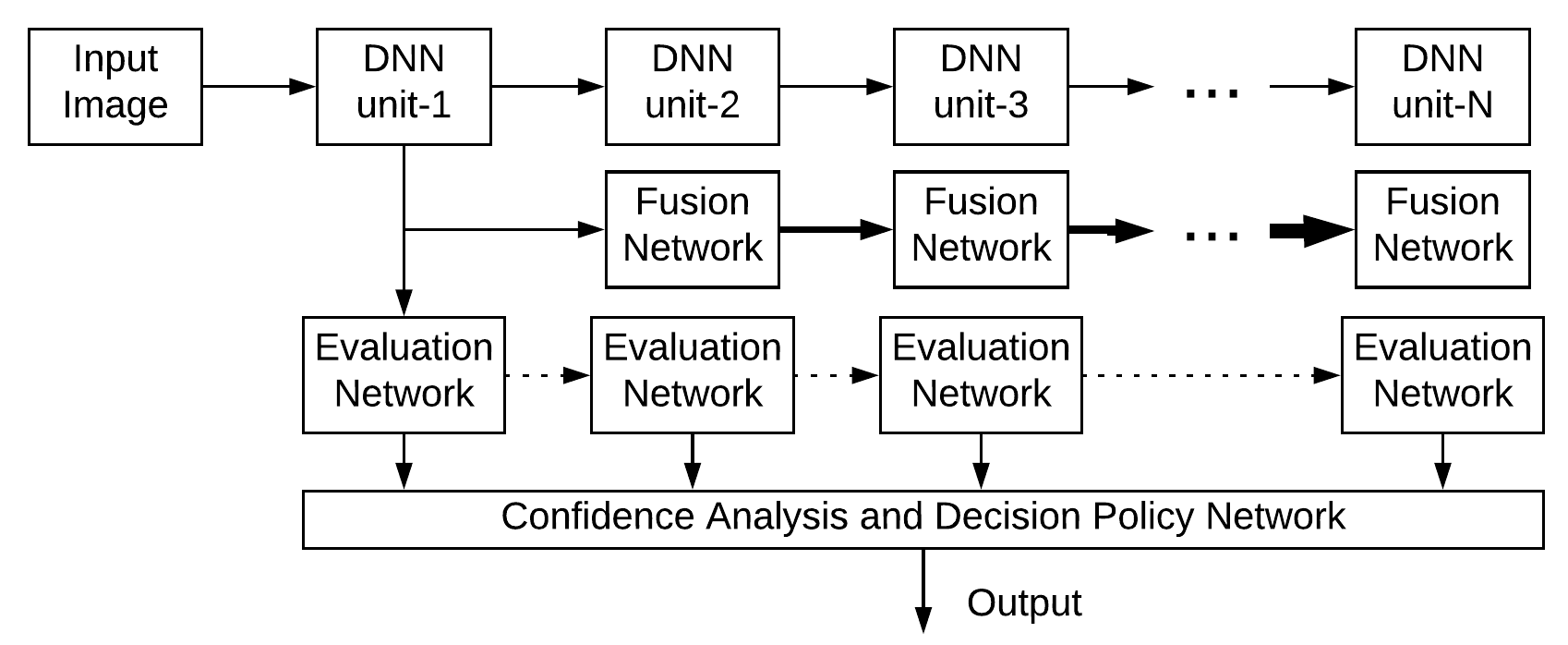}
\caption{\small The proposed framework for progressive deep neural networks. }
\label{fig:concept}
\end{figure}

Fig. \ref{fig:concept} provides an overview of the proposed framework for deep neural networks.  In this work, we divide network into $N$ stages with $N$ deep neural network units, $U_n$, $1\le n\le N$. Each unit $U_n$  consists of a set of  network layers, including convolution, pooling, ReLU, etc. The output of unit $U_n$ is a feature vector $F_n$. At decision stage $1$, we use the feature $F_1$ to generate the decision output, i.e., the classification result for the input image, using an evaluation network $E_n$. The evaluation network consists of a set of network layers, including convolution, pooling, fully connected layers, and softmax layers. At stage $2$,  feature $F_1$ generated from unit $U_1$ and  feature $F_2$ generated from unit $S_2$ are fused together using a fusion network to produce the fused feature ${\cal F}_2$, which will be used by the evaluation network $E_2$ to produce the decision output. The fusion network  consists of feature concatenating  followed by convolution layers for normalization. The fused feature  ${\cal F}_2$ produced at stage 2 will be forwarded to stage $S_3$. At stage $N$, the network unit $S_N$ will produce feature $F_N$, which will be fused with ${\cal F}_{N-1}$ from previous layers to produce the fused feature ${\cal F}_{N}$.  We can see that the proposed progressive deep neural network is able to accumulate and fuse more and more visual features, scale up its visual recognition power by activating more network units and certainly consuming more computational resources. 
Certainly, the network inference can be terminated at any stage. 

Let $O_n$ be the output result produced by the evaluation network $E_n$. Certainly, at later stages, the progressive network is able to accumulate more visual features or evidences for classification, its classification accuracy or visual recognition power will be higher, and the uncertainty for decision will be lower. So, we need a carefully designed module, called 
{\it Confidence Analysis and Decision Policy (CADP)} network to analyze the output results $O_n$ from each stage and its previous stage. It will decide if the current decision is reliable enough with early termination of the inference process or we need to proceed to the next stage to gather more visual evidence. In this work, the CADP network is realized by a recurrent neural network (RNN) learned from the training data.  

 The task the CADP network is two-fold: (1) it needs to generate the decision of classification 
 or other visual recognition tasks at the current stage $S_n$ using the evaluation results 
 $\{O_m| 1\le m\le n\}$ from the current and previous stages. (2) It needs to learn an optimal decision policy for early termination such that the overall computational complexity is minimized while maintaining the state-of-the-art classification accuracy achieved by existing non-progressive networks. 
 Let $x_k$, $1\le k\le K$ be the input image. 
We define 
\begin{align}
E(x_k, n) = \left\{
\begin{array}{ll}
1, & \ \ x_k \text{ is correctly classified at stage}\ n, \\
0, & \ \ x_k \text{ is wrongly classified at stage}\  n. 
\end{array}
\right.
\end{align} 
We denote the decision policy in the CADP network by $\pi[\cdot]$.
The CADP network decides that image $x_k$ be terminated at stage $\pi[x_k]$. 
Let $C(n)$ be the computational complexity of stage $S_n$. 
Then the computational complexity for evaluating the input image $x_k$ will be 
\begin{equation}
C(x_k, \pi) = \sum\limits_{m=1}^{\pi[x_k]} C(m).
\end{equation}
The overall accuracy for all test images will be given by
 \begin{align}
 P(\pi) = \frac{1}{K}\cdot \sum\limits_{k=1}^K E(x_k, \pi[x_k]).
 \end{align}
Therefore, the optimal policy to be learned by the  CADP network aims to minimize the overall complexity while maintaining the target accuracy:
\begin{equation}
\pi = \arg\min\limits_{\pi} C(x_k, \pi)= \sum\limits_{m=1}^{\pi[x_k]} C(m), \\ s.t.\ \ 
P(\pi) = \frac{1}{K}\cdot \sum\limits_{k=1}^K E(x_k, \pi[x_k])\ge P_T.
\end{equation}
In the following sections, we will present our progressive deep neural network design and explain our method to learn the CADP network. 

\section{Progressive Deep Neural Network Design}
\label{sec:progressive}

The concept of progressive inference is different from traditional network inference. It must be able to  produce a sequence of complete prediction results. Early stage of the network should have  small computational complexity. Besides, the features and results from previous stages should be reused and accumulated. 
As discussed in  \cite{sandler2018inverted},  the overall computation required by standard convolutional layers in a CNN is given by:
\begin{align}
  C = \sum_{i=0}^{L}{K_i^2 \times M_i \times N_i \times F_i^2}, \label{eq:cost}
\end{align}
where $K_i$, $F_i$, $M$ and $N$ are kernel size, input feature map size, input and output channels of layer $i$, respectively. To change the values of $K_i$ and $F_i$, we can choose different building blocks, such as  \textit{residual} \cite{he2016deep} and \textit{dense} \cite{huang2017densely}. In complexity-progressive network design, we focus on the rest two sets of parameters: \textit{channels}($M$,$N$) and \textit{layers}(L), which dominate the overall complexity since their values are typically very large. 

It can be seen that these two sets of parameters represent two different dimensions,   corresponding to two different dimensions for network partition: horizontal and vertical partitions. This leads to two different structures, parallel and serial structures, for progress deep neural network design, which will be explained in the following.

\begin{figure}
\centering
\includegraphics[width=0.8\linewidth]{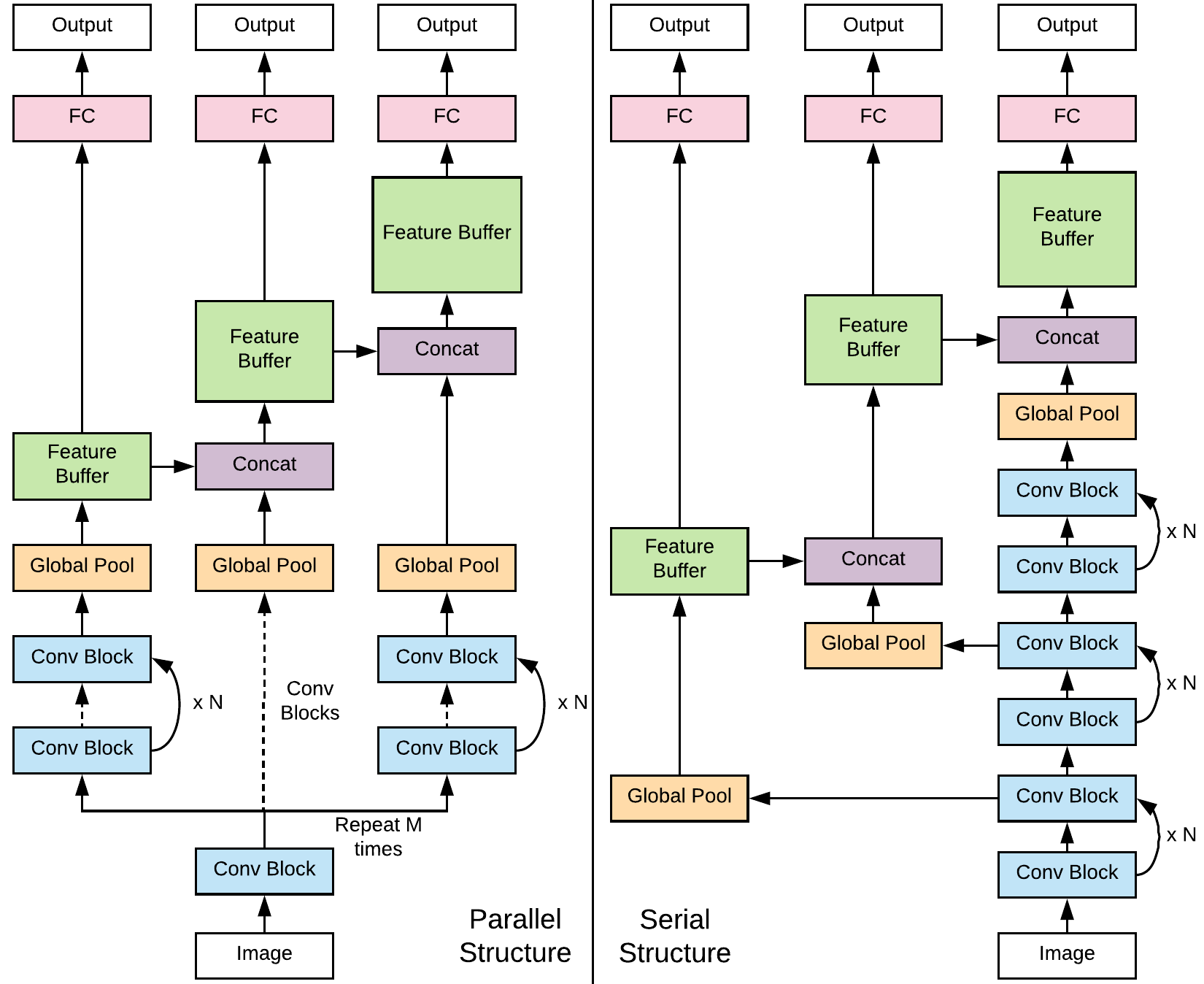}
\caption{\small ProgNet prototype structures of parallel(left) and serial(right) for image classification. Spatial reduction cells are omitted in figures for simplicity.}
\label{fig:parallel_serial}
\end{figure}

\textbf{(A) Parallel structures.}
In the parallel structure for deep neural networks, we partition the network into multiple stages
by reducing the input and output channel size, $M$ and $N$. 
let $r$ be the down-sampling ratio of  $M$ and $N$. The complexity of convolution layers can be then reduced by  $r^2$ according to  Eq. (\ref{eq:cost}). As shown in Fig. \ref{fig:parallel_serial} (left), 
at stage $S_n$, we  use a thin network with small input and output channel sizes. The depth of the network $L$ could be as large as the original non-progressive network. 
We use existing building blocks developed in the literature. Available choices are residual \cite{he2016deep}, residual bottleneck \cite{he2016identity}, dense \cite{huang2017densely}, inception \cite{szegedy2016rethinking}, inception-residual \cite{szegedy2017inception} and NasNet \cite{zoph2017learning} blocks.
Similar to \cite{zoph2016neural}, a \textit{Reduction block} contains stride 2 convolution or down-sampling layers used to reduce spatial size by a factor of 2. A \textit{normal block} keeps spatial dimension intact. In this parallel structure, the input image is analyzed by different network units with different channel sizes. The features generated by different units are fused and accumulated together using a concatenate operator.

\textbf{(B) Serial structures.}
One limitation of the parallel structure is that the width of each unit or branch cannot be reduced to arbitrary numbers. In our experiments, 4 and 8 are the minimum effective width of each unit on the CIFAR \cite{krizhevsky2009learning} and ImageNet \cite{ILSVRC15} datasets, respectively, in order to maintain sufficient representation capacity. 
The serial structure partitions the network along the dimension of layers $L$. 
As shown in Fig. \ref{fig:parallel_serial} (right), we extract features from different layers of the network, apply global pooling to them, and use a fully connected layer to generate the output decision. Also, this feature is concatenated with features extracted from next layers to be used for decision at the next stage. We can see that in this serial structure, the complexity of different stages increases linearly with the layer depth $L$.

Designing and successfully training the progressive network structure is a challenging task. 
Specifically, we need to make sure: (1) the overall accuracy at stage 
$S_n$ achieved by the evaluation network $E_n$  is increasing with $n$. Otherwise, additional computational computational resources have been wasted. (2) When we apply the full complexity, i.e., evaluate each input using the whole network, we need to make sure that it is more complexity-accuracy effective than existing state-of-the-art networks.  

Following the work in \cite{krizhevsky2012imagenet,simonyan2014very} for multi-class classification, we use the  Cross Entropy loss  as our joint loss function:
\begin{align}
  \mathcal{L} = \sum\limits_{m=1}^{M}{w_m \cdot \mathcal{L}_{CrossEntropy}(y_m, \hat{y})}
\end{align}
where $w_m$ and $y_m$ are weight and output from stage $m$, respectively. $\hat{y}$ is the ground-truth label. If not otherwise required by targeting application, we treat each stage with uniform weights ($\vec{w} = \vec{u}$), \ie outputs from all stages are equally important.

\section{Confidence Analysis and Decision Policy Learning}
\label{sec-cadp}

\begin{figure}
\centering
\includegraphics[width=0.6\linewidth]{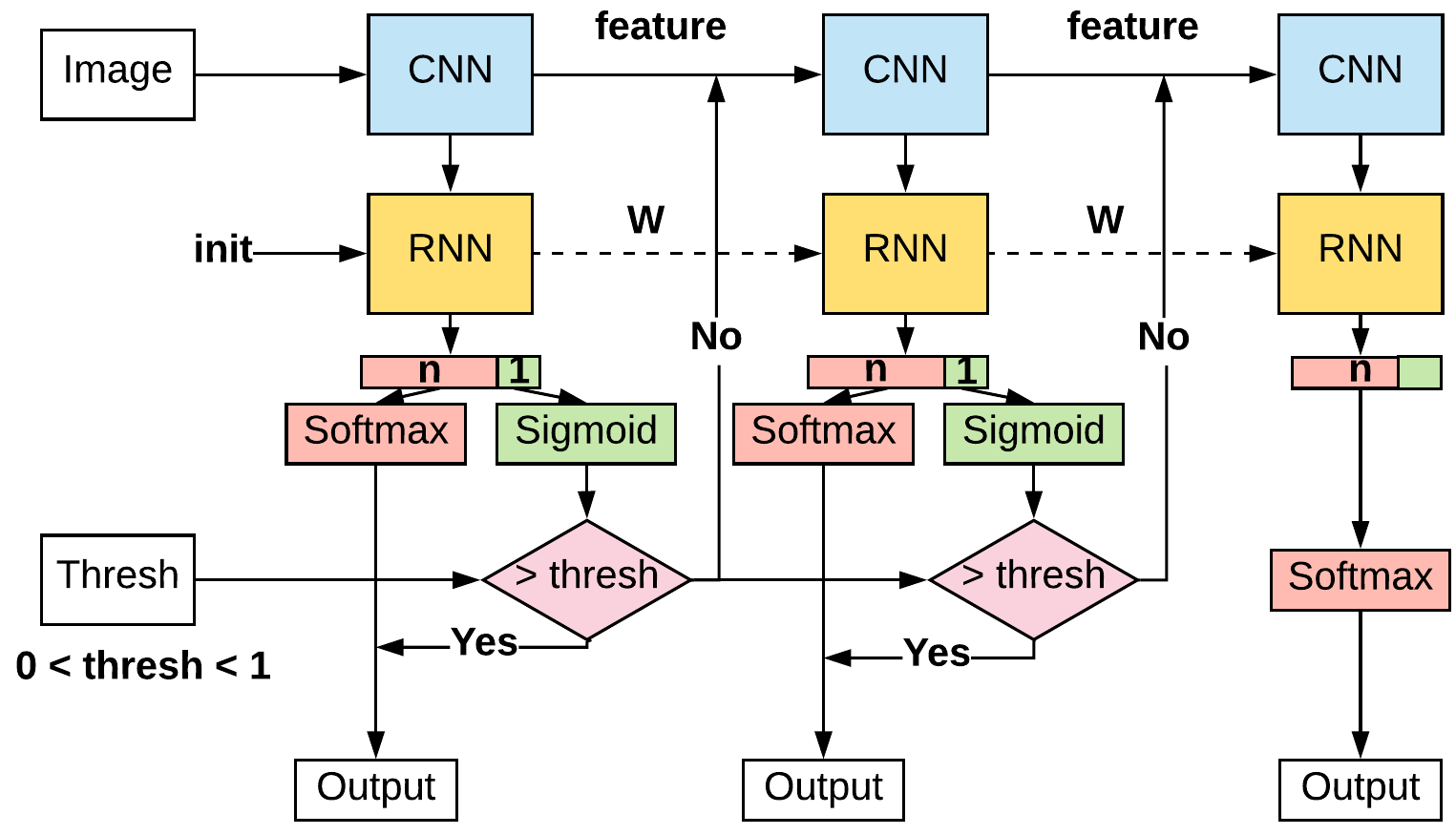}
\caption{\small Early termination control during inference by RNN controller. }
\label{fig:controller}
\end{figure}

In the previous section, we have introduced the ProgNet  that can perform network prediction at a sequence of stages. At each stage, ProgNet needs to determine if the current evaluation output is reliable or confident enough and if it is necessary to proceed to the next stage for accumulate more visual evidences. During our experiments, we found out that the decisions at different stages are inter-dependent with each other. Specifically, the current stage needs to examine the evaluation results in all previous stages for effective decisions.  To address issue, we propose to design 
a recurrent neural network (RNN) to learn the confidence analysis and decision policy (CADP). 
As illustrated in Fig. \ref{fig:controller}, the RNN CADP network uses the \textit{pre-softmax} outputs in the evaluation networks of all previous stages to learn the confidence estimation and decision policy at the current stage. 

More specifically, for each input (usually a mini-batch of images), a RNN controller takes as input the $n$-class \textit{pre-softmax} output from the current CNN classifier, and produces $n + 1$ outputs, with  $n$ new categorization results  and $1$ \textit{post-sigmoid} confidence estimation. \textit{Post-sigmoid} confidence score is compared with user defined threshold $t$ in range $(0, 1)$ to determine whether output is emitted directly from RNN classification results, otherwise another stage of CNN-RNN-decision is required.

Optimizing the RNN controller is the most challenging problem in this work. For each image, with a user-defined threshold $t \in (0, 1)$, the objective of CADP network is to solve the following optimization problem: minimizing the overall error rate subjected to a computational complexity constraint: 
\begin{align}
 & \underset{x}{\text{minimize}} & & \mathcal{L}_{CADP} = \int_{0}^{1}{\sum_{i=1}^{M}{c_i \cdot \prod_{j=1}^{i}{H(z_j - t)}}} dt + \lambda \cdot \int_{0}^{1}{(1 - \hat{y_p})} dt, \\
& & &  p = \argfind_{i=0}^{M-1}{[\text{where}(H(z_i - t) = 1 \land H(z_{i+1} - t) = 0 )]}, \label{eq:find_p} \\
 & \text{subject to} & & 0 < z_i < 1, \qquad z_i \in \mathbb{R}  \label{eq:constraint}
\end{align}
where $M$ is the number of stages and $C_i$ is the normalized computational cost of $i$-th DNN  unit. $C_i$ is a constant number once ProgNet is composed. $H(\cdot)$ is Heaviside(unit) step function which is $1$ for positive inputs and $0$ for negative inputs. $\hat{y_p}$ is the correctness score ($1$ if correctly classified, $0$ otherwise) of the $p$-th stage which  is the last rejected stage by $t$. Finding $p$ is  equivalent to finding the first index where $(H(z_i - t) = 1)$ and $(H(z_{i+1} - t) = 0)$ as shown in Eq. (\ref{eq:find_p}). 
Without loss of generality,  we approximate the confidence integral $\int_0^1{(\cdot) dt}$ using  summation over discrete samples of $T$ within the range of  $(0, 1)$.  The above optimization problem becomes :
\begin{align}
  \mathcal{L}_{conf} = \sum_{t \in T}{\sum_{i=0}^{M}{c_i \cdot \prod_{j=0}^{i}{H(z_j - t)}}} + \lambda \cdot \sum_{t \in T}{(1 - \hat{y_p})}. \label{eq:l_conf}
\end{align}
Combining standard cross entropy loss and $\mathcal{L}_{conf}$, we have the following optimization objective function:
\begin{align}
  & \underset{\mathbf{w^*}, \mathbf{b^*}}{\text{minimize}}   & -\sum{y_k \cdot \log{\hat{y_k}}} + \alpha \cdot \sum_{t \in T}{\sum_{i=0}^{M}{c_i \cdot \prod_{j=0}^{i}{H(z_j - t)}}} + \lambda \cdot \sum_{t \in T}{(1 - \hat{y_p})}, \label{eq:l_controller} 
\end{align}
where $\mathbf{w^*}$ and $\mathbf{b^*}$ are weights and bias of the RNN controller, respectively. $y_k$ is the ground-truth of $k$-th image, $\alpha$ is the hyper-parameter controlling weights of classification and confidence losses. While it is  possible to directly optimize  Eq. (\ref{eq:l_controller}) with constraints Eq. (\ref{eq:constraint}) using the method in  \cite{pathak2015constrained}, we found that it is more efficient and robust to convert the problem into:
\begin{align}
  & \underset{z_j \in \mathbf{z^*}}{\text{minimize}} &  \sum_{t \in T}{\sum_{i=0}^{M}{c_i \cdot \prod_{j=0}^{i}{H(z_j - t)}}} + \lambda \cdot \sum_{t \in T}{(1 - \hat{y_p})}, \label{eq:minimize_z} \\
  & \underset{\mathbf{w^*}, \mathbf{b^*}}{\text{minimize}} & -\sum{y_k \cdot \log{\hat{y_k}}} + \alpha \cdot {\sum_{i=0}^{M}{|z_i - z_i^*|}}, \qquad \label{eq:sgd}
\end{align}
where $z^*$ is the  optimal confidence score. The problem in Eq. (\ref{eq:minimize_z})can be solved with  the Constrained Optimization by Linear Approximation algorithm \cite{powell2007view}.  Eq. (\ref{eq:sgd}) can be solved  using back-propagation with a standard stochastic gradient descend (SGD).

Note that the desired $\hat{y_p}$ in Eqs. (\ref{eq:l_conf}), (\ref{eq:l_controller}), and  (\ref{eq:minimize_z}) is the output from the RNN classifier, while it is possible to update $\hat{y_p}$ after each batch or epoch, it is a very expensive process. In this work we are using the outputs from the evaluation networks for the first $80 \%$ controller training epochs. We then update using the latest $\hat{y_p}$ from the controller output, and continue training controller for the rest $20 \%$ epochs.  We implement the RNN controller using a three-layer LSTM, stacking 3 LSTM cells with 2 Dropout \cite{srivastava2014dropout} layers in between. Before each forwarding step, internal states of controller are initialized with zero inputs. More training details will be provided in in Section \ref{sec:results}.

\section{Experimental Results}
\label{sec:results}

In this section, we evaluate our proposed ProgNet using the CIFAR-10 \cite{krizhevsky2009learning} and ImageNet(ILSVRC2012) \cite{ILSVRC15} datasets. On both datasets, our goal is to train a single ProgNet 
model which provides progressive complexity-accuracy scability while outperforming existing state-of-the-art networks in terms of complexity-accuracy performance. 
All ProgNet models are trained on AWS P3 8x large instances \footnote{https://aws.amazon.com/cn/ec2/instance-types/p3/} with 4 Tesla V100 GPUs. Testings and run-time benchmarks are executed on local workstation with 1 Intel Xeon(R) CPU E5-1620 v3 @ 3.50GHz and 4 Pascal Titan GPUs. We implement the ProgNet using Gluon \footnote{https://mxnet.incubator.apache.org/tutorials/gluon/gluon.html} imperative python APIs with MXNet backend \cite{chen2015mxnet}. All  reference networks for performance comparison are also benchmarked using MXNet if not otherwise specified.

\textbf{(A) Network Configurations.}
Both parallel and serial structures of the ProgNet are flexible and highly expandable. In this work, we conducted extensive experiments using three different network settings for CIFAR-10 and two different settings for ImageNet,
which are summarized in Table \ref{tab:prognets}.

\setlength{\tabcolsep}{4pt}
\begin{table}
\centering
\caption{Network Configurations for CIFAR-10 and ImageNet Datasets.}
\label{tab:prognets}
\begin{adjustbox}{width=1\textwidth}
\begin{tabular}{c|c|c|c}
\specialrule{.1em}{.05em}{.05em} 
\multicolumn{4}{c}{CIFAR-10} \\
\specialrule{.1em}{.05em}{.05em} 
Multiplier & {[}1, 1, 2, 3{]} & {[}1, 1, 1, 2, 3, 4{]} & {[}1{]} \\ \hline
Output & ProgNet-p4-residual & ProgNet-p6-residual & ProgNet-s9-dense, k=12   \\ \hline
$32\times 32$     & $3\times 3$ conv, stride 2                               & $3\times 3$ conv, stride 2                                     & $3\times 3$ conv, stride 2 \\
$16\times 16$     & {[}res $\times$ 2{]} - {[}$3\times 3$ s2 max pool{]}        & {[}res $\times$ 2{]} - {[}$3\times 3$ s2 max pool{]} & {[}d $\times$ 2, fc{] $\times 3$} - {[}$3\times 3$ s2 max pool{]}                 \\
$8\times 8$      & {[}res $\times$ 3{]} - {[}$3\times 3$ s2 max pool{]}                               & {[}res $\times$ 3{]} - {[}$3\times 3$ s2 max pool{]}           & {[}d $\times$ 2, fc{] $\times 3$} - {[}$3\times 3$ s2 max pool{]}                  \\
$4\times 4$      &    {[}res $\times$ 3{]} - {[}global avg pool{]}                               & {[}res $\times$ 3{]} - {[}global avg pool{]}          &  {[}d $\times$ 3, fc{] $\times 2$}  - {[}global avg pool{]}  \\
$1\times 1$  &  fc $\times 4$ & fc $\times 6$ & fc \\
\hline              
\end{tabular}
\end{adjustbox}
\vspace{6em}
\begin{adjustbox}{width=1\textwidth}
\begin{tabular}{c|c|c}
\specialrule{.1em}{.05em}{.05em} 
\multicolumn{3}{c}{ImageNet} \\
\specialrule{.1em}{.05em}{.05em} 
Multiplier & {[}1, 1, 2, 3{]} &  {[}1{]} \\ \hline
Output & ProgNet-p4-residual & ProgNet-s6-dense, k=18                              \\ \hline
$112\times 112$    & $7\times 7$ conv, stride 2                               & $7\times 7$ conv, stride 2                            \\
$56\times 56$     & $3\times 3$ max pool, stride 2                           & $3\times 3$ max pool, stride 2                        \\
$28\times 28$     & {[}res $\times 2${]}- {[}res, stride 2{]}                              & {[}d $\times 7${]}-{[}d, stride 2, fc{]}  \\
$14\times 14$     &   {[}res $\times 3${]} - {[}res, stride 2{]}    & {[}d $\times 11${]}-{[}d, stride 2, fc{]} \\
$7\times 7$      &     {[}res $\times 4${]}-  {[}global avg pool{]}         & {[}d $\times 12$, fc, d $\times 8$, fc, d $\times 6$ , fc, d $\times 4${]}- {[}global avg pool{]}\\ 
$1\times 1$  &  fc $\times 4$ & fc \\
\hline
\end{tabular}
\end{adjustbox}
\vspace{-6em}
\end{table}
\setlength{\tabcolsep}{1.4pt}

\textbf{(B) Network Training and Inference.}
The base classifier and LSTM controller in ProgNet are trained separately using SGD optimization. For the base network, we use a  batch size of 256. The number of training epochs for CIFAR-10 and ImageNet are 350 and 180, respectively. Following \cite{huang2017densely,zoph2016neural,zoph2017learning}, we use an initial learning rate 0.1, weight decay 0.0001, and momentum 0.9. The learning rate is lowered by a factor of 10 at 25\%, 50\% and 80\% of total epochs. Parameters with the best mean accuracy of all DNN units are saved as our best model. We then start training the LSTM controller using this best model. We sample the early termination threshold  $T = \{x \times 0.1 \mid x = 1, 2, ..., 9\}$. $0$ and $1.0$ are excluded because $0$ are considered as no network inference and $1.0$ stands for full network  inference. The controllers are optimized using SGD with a learning rate 0.5, weight decay 0 and momentum 0 for all experiments. 

To evaluate the impact of RNN controllers during inference, we conducted  experiments using the following two modes: \textit{dynamic} and \textit{fixed}.
In the \textit{dynamic} mode, users can specify the confidence $T$ as early termination threshold. This is the desired behavior in this work. 
For comparison, we allow the ProgNet to follow preset stage setting and run in \textit{fixed} mode. Once set, The ProgNet acts as a  non-progressive network.

In our experiments, we evaluate two different progressive structure: parallel and serial, with different stages, such as 4, 6, and 9 stages. We also evaluate two different modes, dynamic and fixed, for the CADP controller.
For example, in our figures, the notation \textit{p6-dynamic} indicates that the network has a parallel structure with 6 stages and a dynamic CADP mode. \textit{s9-fixed} indicates that the network has a serial structure with 9 stages and a fixed CADP mode.

\subsection{Experimental Results on the CIFAR-10 Dataset}
\begin{figure}[tb]
\centering
\includegraphics[width=.95\linewidth]{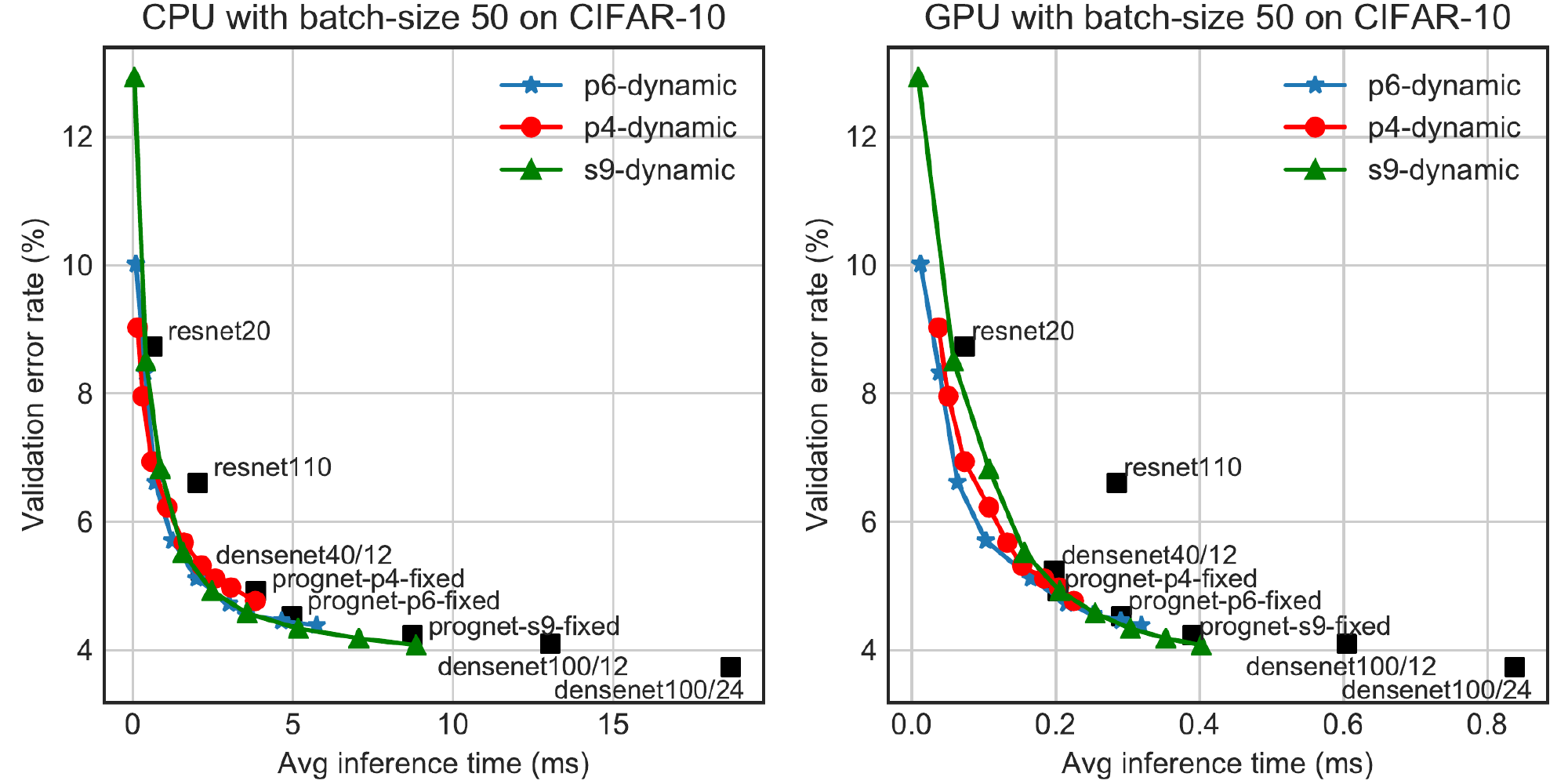}
\caption{\small Error rate versus actual average inference speed on CIFAR-10 validation set. Black squares indicate results from  previously published networks running on our machine.}
\label{fig:cifar-speed}
\end{figure}

In our CIFAR-10 experiments, we follow previous practice \cite{he2016deep} to augment the  data. We pre-process the training images by converting them to \textit{float}32, followed by up-sampling and random cropping to $32 \times 32$. Random horizontal flipping is applied during training as a common strategy. The validation data is processed by converting to \textit{float}32 without augmentation.
We set the batch size to be 50. 
For a  given early termination threshold $T$, the confidence analysis and decision policy network in ProgNet 
will decide when to stop the network inference for the input image. Let $C(T)$ be the corresponding average network inference time of ProgNet. We also record the overall error rate by $P(T)$. Fig. \ref{fig:cifar-speed} shows the $P(T)$ curves the following network settings: \textit{p6-dynamic, p4-dynamic, and s9-dynamic}. 
We shows the results on CPU (left) and GPU (right).
We include results for ProgNet with fixed controller modes (shown in solid boxes). For comparison, 
we also include the complexity-accuracy results for  state-of-the-art networks on the CIFAR-10 dataset: \textit{resnet20, resnet10, densenet100/12, densetnet100/24}.
It can be seen that our proposed ProgNet outperforms existing networks in the complexity-accuracy performance using one single model. As we increase the number of stages, we can achieve a large complexity-accuracy scalability range. The parallel structure is slightly better than the serial structure.

\begin{figure}
\centering
\hspace{-0.7cm}\parbox{5cm}{
\includegraphics[width=1.15\linewidth]{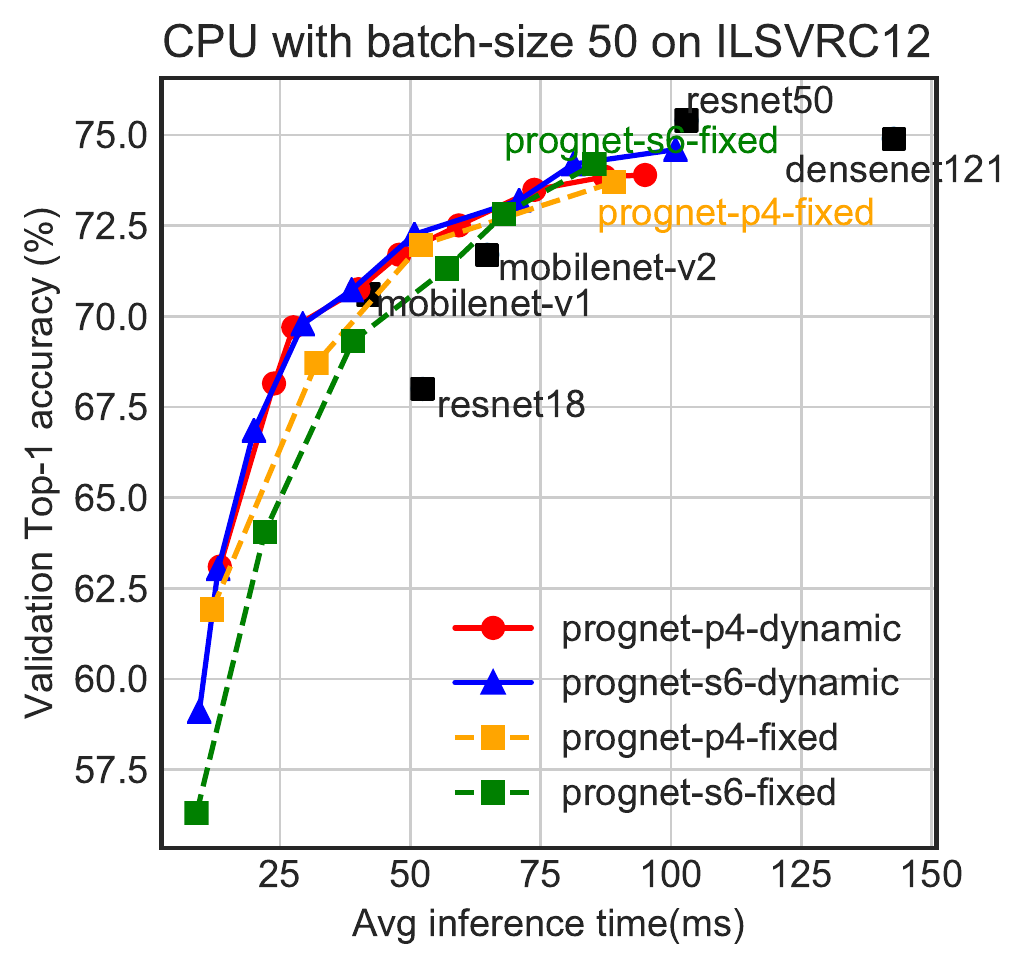}}
\qquad
\begin{minipage}{5cm}
\includegraphics[width=1.15\linewidth]{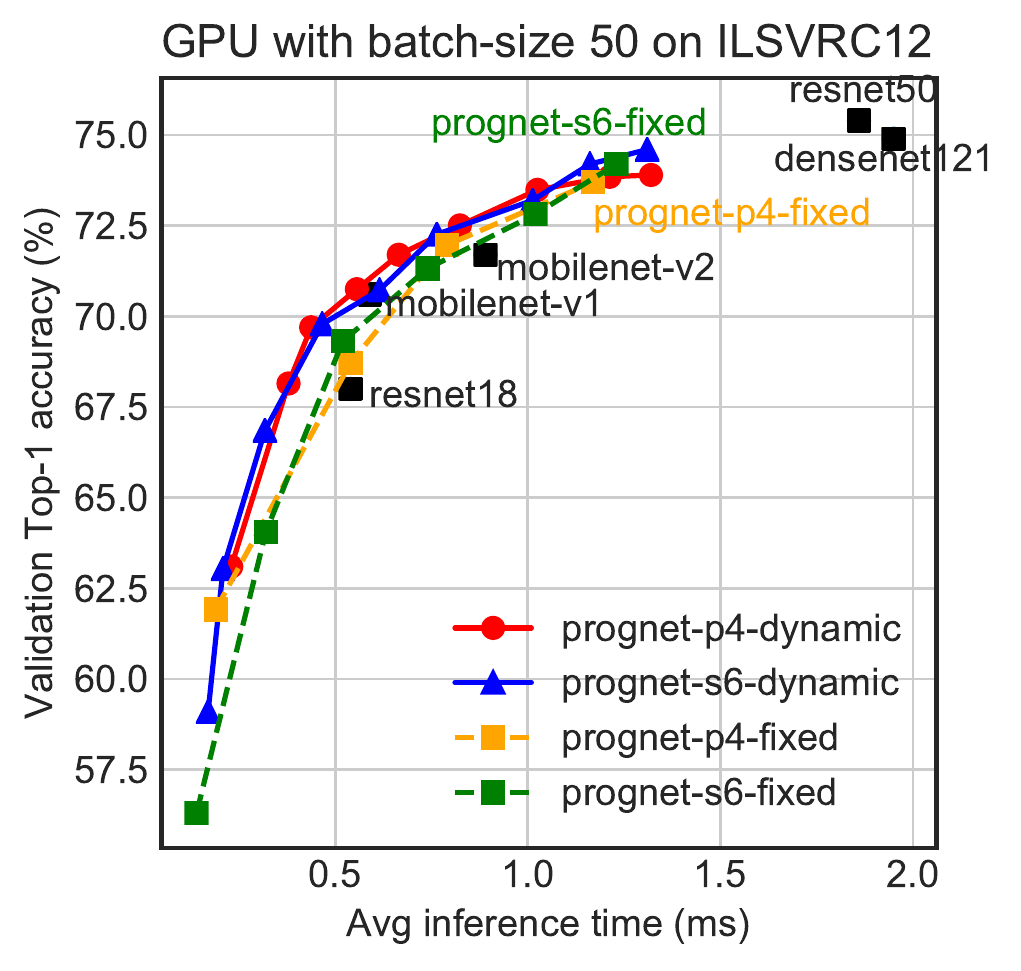}
\end{minipage}
\caption{\small Top-1 accuracy versus actual average inference speed on the ImageNet validation set. Black squares represent results from  previously published networks. }
\label{fig:imagenet-speed}
\end{figure}

\subsection{ Experimental Results  on the ImageNet}
Following the same procedures outlined in existing papers \cite{he2016identity,huang2017densely,zhang2017shufflenet,zoph2016neural,zoph2017learning}, we conduct the experiments of ProgNet training and  testing on the  ILSVRC 2012  \cite{ILSVRC15}. We use an image size $224 \times 224$. Training images are augmented with random  cropping with min/max aspect ratio of $3 / 4$ and $4 / 3$. Random horizontal flipping is used in training. For training and validation, pixel means of [123, 117, 104] are subtracted from images and then normalized by standard deviations of [58.4, 57.12, 57.38]. As in \cite{he2016deep,he2016identity,zoph2016neural}, we use 1.28 million images for training and 50000 images for testing.

Similar to the above CIFAR-10 experiments, we record the complexity-accuracy curve $P(T)$ for different ProgNet structures (parallel and serial) with different stages (4, and 6 stages) using different controller modes (dynamic and fixed). Fig. \ref{fig:imagenet-speed} shows these curves for CPU (left) and GPU (right). We also include complexity-accuracy results achieved by existing networks, including \textit{resnet18, resnet50, densenet121, mobilenet-v1, mobilenet-v2}. 
Table \ref{tab:imagenet-perf} summarizes the complexity-accuracy performance of ProgNet in comparison with existing networks. We  include results on the number of model parameters, number of MACC (Multiply-Accumulation) operations, and running times on CPU and GPU. 
It can be seen that our proposed ProgNet outperforms existing networks in the complexity-accuracy performance by a large margin using one single model. 
For the same complexity, both ProgNet variants outperform the MobileNet  \cite{howard2017mobilenets,sandler2018inverted}, which has been significantly optimized,
by  2\% to 3\% in classification accuracy.   For the same accuracy, the ProgNet-p4-dynamic model is able to achieve 20\% less inference time than  MobileNet-v2.

\setlength{\tabcolsep}{4pt}
\begin{table}[]
\centering
\caption{\small Complexity-accuracy comparison between ProgNet and existing networks on the ImageNet dataset.  }
\label{tab:imagenet-perf}
\begin{adjustbox}{width=0.7\textwidth}
\begin{tabular}{c|c|cc|cc}
\specialrule{.1em}{.05em}{.05em} 
Model            & Top-1 & Params & MACC  & CPU & GPU \\ \specialrule{.1em}{.05em}{.05em} 
ResNet-18 \cite{he2016deep}       & 68.0  & 11.69M & 1.83G & 52.37ms  &  0.54ms   \\
ResNet-50 \cite{he2016deep}       & \textbf{75.4}  & 25.26M & 3.87G & 103.00ms  &  1.86ms   \\
DenseNet-121 \cite{huang2017densely}     & 74.9  & 7.98M  & 3.08G & 142.73ms  &  1.95ms   \\
MobileNet-v1 \cite{howard2017mobilenets}     & 70.6  & 4.2M   & 575M  & 41.98ms  &  0.59ms   \\
MobileNet-v2 \cite{sandler2018inverted}    & 71.7  & 3.4M   & 300M  & 64.78ms  &  0.89ms   \\
ShuffleNet(1.5) \cite{zhang2017shufflenet} & 69.0  & 2.9M   & 292M  &  -  &  -  \\
NasNet-A \cite{zoph2017learning}        & 74.0  & 5.3M   & 564M  &  -  &  -  \\ 
\specialrule{.1em}{.05em}{.05em} 
\specialrule{.1em}{.05em}{.05em} 
ProgNet-p4-dynamic &  73.9  &  13.12M   & 1.87G   & 89.03ms  & 1.17ms  \\ 
ProgNet-s6-dynamic & 74.6  & 14.3M   & 1.31G & 85.31ms  &  1.23ms  \\
\hline
ProgNet-p4-fixed @t=0.5 & 71.9 & 13.12M & - & 47.3ms & 0.67ms \\
ProgNet-p4-fixed @t=0.6 & 72.5 & 13.12M & - & 59.8ms & 0.81ms \\
ProgNet-s6-fixed @t=0.3 & 67.1 & 14.3M & - & \textbf{23.2ms} & \textbf{0.31ms} \\
\hline
\end{tabular}
\end{adjustbox}
\end{table}
\setlength{\tabcolsep}{1.4pt}

\begin{figure}
 \centering
 \includegraphics[width=0.45\linewidth]{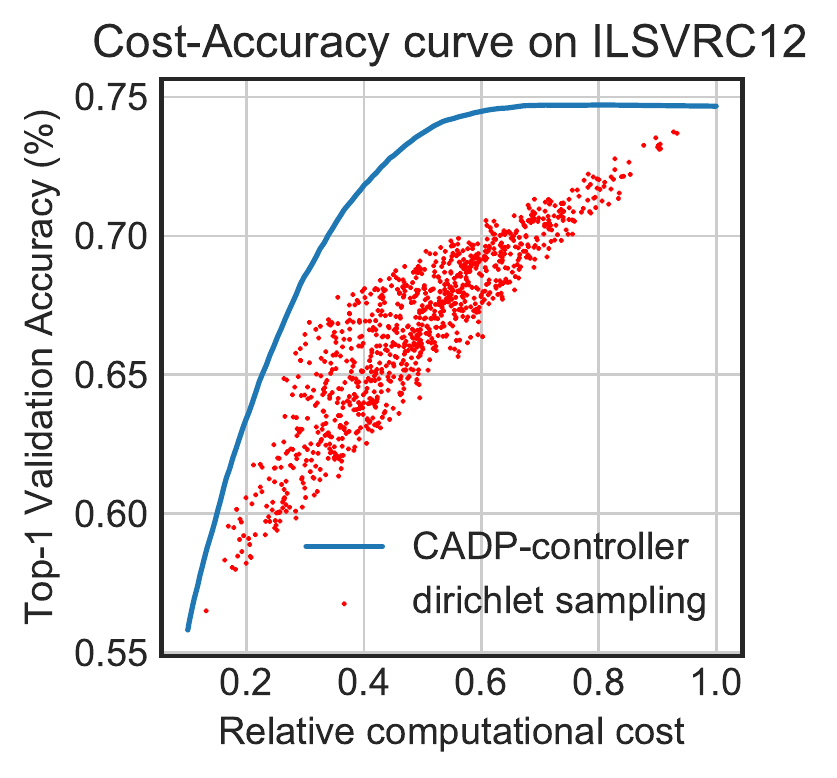}
 \vspace{-0.5cm}
 \caption{Complexity-accuracy performance comparison between our CADP controller and random termination.}
 \label{fig:cost-accuracy}
 \end{figure}

In our ProgNet design, the confidence analysis and decision policy (CADP) network plays a critical role since it controls whether the next network stage should be activated for the input image or its inference should be early terminated. This has a direct impact on the complexity-accuracy performance of our ProgNet.  To further understand the behavior and capability of the CADP controller, we implement a random controller in which the network inference of an input image is terminated at a random stage. We then run the experiment using this random controller on the ImageNet dataset for 1000 times to simulate partial brute-force search for the best control policy. In Fig. \ref{fig:cost-accuracy}, one dot represents the complexity-accuracy result for one experimental run. The solid curve represents our CADP controller. We can see that the CADP is very effective, outperforming static control. 

\section{Conclusion}
\label{sec:conclusion}
In this work, we have successfully established a progressive deep neural network structure 
which is able to adapt its network inference structure and computational complexity to images with different visual recognition complexity. This progressive structure is able to scale up its discriminative power and achieve higher recognition capability by activating and executing more analysis units of the deep neural networks  for more difficult vision analysis tasks or image sets.  
We have developed a multi-stage progressive structure, called ProgNet,  for deep neural networks, with each stage being separately evaluated for output decision and all stages being  activated in a sequential manner with progressively increased complexity and visual recognition power.
 We developed a recurrent neural network method to learn  the confidence analysis and decision policy for early termination. Our extensive experimental results on the CIFAR-10  and ImageNet datasets demonstrated that the proposed ProgNet is able to obtain more than 10 fold complexity scalability while achieving the state-of-the-art performances with a single network model. 

\clearpage

\bibliographystyle{splncs}
\bibliography{egbib}
\end{document}